# Occlusion Robustness of CLIP for Military Vehicle Classification

Jan Erik van Woerden[*], Gertjan Burghouts, Lotte Nijskens, Alma M. Liezenga, Sabina van Rooij, Frank Ruis, and Hugo J. Kuijf

TNO, Oude Waalsdorperweg 63, 2597 AK, the Hague, the Netherlands

## ABSTRACT

Vision-language models (VLMs) like CLIP enable zero-shot classification by aligning images and text in a shared embedding space, offering advantages for defense applications where labeled data is scarce. However, CLIP is primarily trained on high-quality internet imagery. Its robustness in challenging military operational environments, characterized by factors like partial occlusion and degraded signal-to-noise ratio (SNR) because of obscurant objects or weather, remains underexplored. We investigate the robustness of several CLIP variants to occlusion, using a custom dataset of 18 military vehicle classes. We simulate both contiguous occlusions (slide blackout, bar occlusion) and dispersed occlusions (random rain, snow, grid dropout) to reflect real-world environmental challenges. Robustness is evaluated using Normalized Area Under the Curve (NAUC) across occlusion percentages.

Four key insights emerge: (1) Fine-grained, dispersed occlusions (e.g., snow, rain) degrade performance more than larger contiguous occlusions—NAUC of 61.3% for dispersed occlusion vs 78.9% for contiguous occlusions for PE-Core-ViT-L/14-336 (2) Transformer-based CLIP models consistently outperform CNN-based CLIP models, with ViT-B/16 achieving up to 22 percentage point higher NAUC than ResNet50; (3) Pre-training methodology significantly affects robustness—PE-Core models consistently outperform CLIPA counterparts at similar scales (e.g., +6.7pp NAUC at ~320M parameters), showing that improved pre-training augments robustness beyond scaling alone; (4) Fine-tuning introduces a trade-off— linear probing boosts clean-image accuracy (55.6%→88.0%) but reduces robustness under dispersed occlusions (snow NAUC 54.0%→36.0%), while full fine-tuning mitigates this effect (snow NAUC 44.5%) yet still falls short of zero-shot consistency. These results underscore the importance of occlusion-specific augmentations during training and the need for further exploration into patch-level sensitivity and architectural resilience for real-world deployment of CLIP.

## INTRODUCTION

Military vehicle classification in operational environments presents unique challenges that are often not present in standard computer vision benchmarks [1] [2]. Defense applications require reliable classification under degraded visual conditions including smoke, dust, foliage occlusion, and varying weather conditions. During reconnaissance and targeting operations, environmental factors can partially or significantly obscure vehicle features while operators still require accurate identification for tactical decision-making. Current military vision systems typically rely on extensive labeled datasets specific to each vehicle type and environmental condition, creating operational limitations when encountering novel vehicles or deployment scenarios.

Recent research has examined CLIP's robustness to various perturbations, primarily focusing on adversarial attacks and common corruptions such as noise and blur [4]. These studies demonstrate that while CLIP shows greater robustness than traditional supervised models to certain distribution shifts, its performance still degrades significantly under targeted perturbations. However, existing robustness evaluations mainly assess uniform corruptions applied across entire images, failing to capture the structured occlusions encountered in real-world scenarios, like foliage. The few studies addressing partial occlusion in vision-language models typically employ simple rectangular masks or random pixel dropout, which poorly represent the complex occlusion patterns created by environmental factors.

---

[*] Please send further correspondence to Jan Erik van Woerden
E-mail: jan_erik.vanwoerden@tno.nl

Furthermore, current evaluation metrics often report only accuracy at fixed occlusion levels, providing limited insight into how models degrade across varying degrees of occlusion. This approach obscures critical performance characteristics, such as whether models fail gradually or exhibit sudden performance drops at specific occlusion thresholds, which is important information for operational deployment decisions. The absence of systematic evaluation using realistic occlusion types and comprehensive metrics leaves significant uncertainty about CLIP's suitability for defense applications where partial visibility is common rather than exceptional.

In this paper we try to solve several of these gaps. We employ the NAUC (Normalized Area Under the Curve) metric to measure robustness and additionally we analyze performance under occluded conditions of a set of CLIP models of different architectures and sizes. We introduce artificial occlusions which both represent contiguous occlusions which can for example represent moving from behind a tree (sliding window), but also dispersed occlusions, such as rain and snow. Using this setup we try to answer the following questions: (i) Do different types of occlusion have a different performance impact on robustness? (ii) How do architectures like CNNs and Transformers differ in robustness in occluded scenarios? (iii) Do different pre-training methodologies impact the robustness of the model? (iv) Do different finetuning techniques impact the final robustness of the model?

## METHOD

To evaluate the abovementioned research questions, we use a private dataset of 18 different military vehicles, which we explain below. This dataset is artificially occluded using 5 different occlusion types: 3 contiguous and 2 dispersed occlusions. We evaluate a set of 9 pre-trained CLIP models varying in architecture, size and training methodology. Lastly, using the metrics accuracy and NAUC, the peak performance and robustness of the different models is evaluated to answer the research questions.

Note: For writing this paper we used Microsoft Copilot to rewrite parts of the text to increase structure and readability. All ideas, experiments and conclusions are our own.

**Dataset**

For this work we continue on the dataset used in [5]. The existing dataset of 13 classes is extended with 5 extra classes to increase difficulty and diversity. This new dataset is compiled from public sources and consists of 18 vehicle classes representing diverse military platforms: tanks (T-62, T-72, T-90, Leopard 2, M1A2), armored personnel carriers (BTR-80, BMP-1, Fuchs, Boxer, Patria), reconnaissance vehicles (BRDM-2, Fennek), self-propelled artillery (2S1, 2S3, M109, MSTA, Panzerhaubitze 2000), and support vehicles (military truck). Some example images are shown in Figure 1. For each class, approximately 50 images were collected; of which 16 were randomly placed in a training set, 4 in a validation set and the remainder in a separate test set.

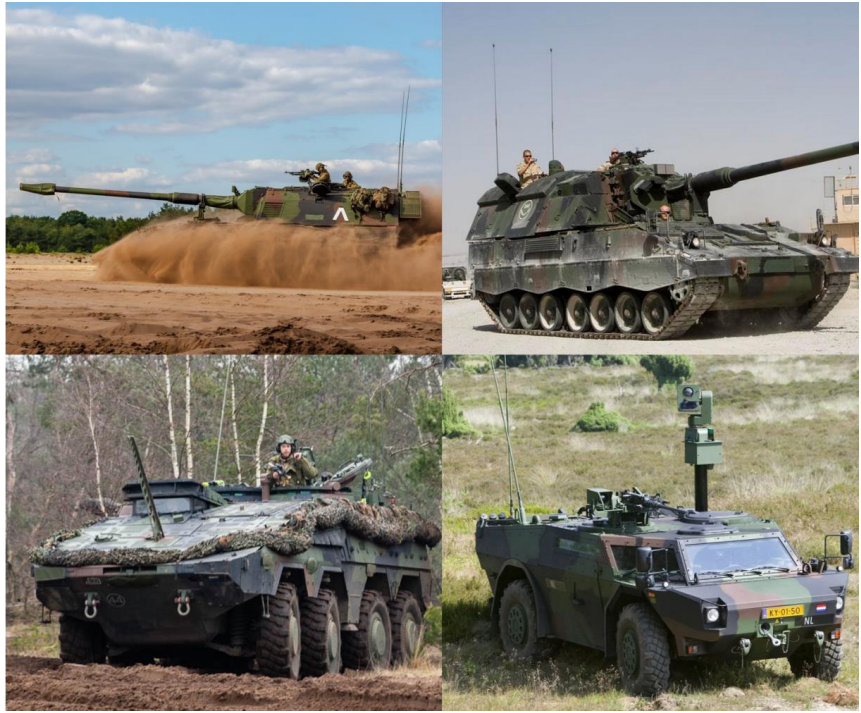

Figure 1: Example images of various military vehicles obtained from public sources. Top row: a Panzerhaubitze 2000 (left image credit: John van den Boogaart, defensiefotografie.nl; right image credit: Netherlands Ministry of Defence); bottom left: a Boxer (image credit: Martin Bos, defensiefotografie.nl); bottom right: a Fennek.

**Models**

We use multiple CLIP architectures to analyze the impact of occluded images on performance. We use models from 4 different training methodologies: (i) the original pre-trained CLIP models [3], (ii) pre-trained models by the CLIPA paper [6] which utilize an improved training methodology and an improved dataset and (iii) a Perception Encoder-Core [7] which further improves on both training methodology and dataset. Lastly (iv) we finetune one of the CLIPA models using our internal linear probing and full finetuning framework for robustness comparison of zero-shot and few-shot learning models.

Training methodologies:

(i) We utilize the original CLIP paper [3] models to compare between Convolutional Neural Network models and Transformers. All models contain roughly the same number of parameters (150-180M). The ResNet-50x4 variant (RN50x4) utilizes a scaled ResNet architecture with 4× wider layers than standard ResNet-50. Two Vision Transformer (ViT) variants complete our baseline selection: ViT-B/32 processes 32×32 image patches with a base transformer architecture, while ViT-B/16 employs finer 16×16 patches for improved spatial resolution.

(ii) The CLIPA models [6] leverage an inverse scaling law for efficient training, whereby the start of the training is done on small training images and further in the training process the images are scaled to the original 224x224 size. This approach reduces computational requirements by 33x compared to traditional CLIP training while achieving competitive accuracy. We evaluate three CLIPA variants with increasing capacity: ViT-L/14 with 14×14 patch size. Larger models were also released, enabled by reduced computational requirements.. ViT-G/14 extends to a larger architecture while maintaining the same patch granularity, and ViT-H/14 represents the largest model with 336×336 input resolution. All models use the Datacomp-1B dataset for training and utilize the standard OpenCLIP architecture [7].

(iii) For the last pre-trained methodology we use Perception Encoder models [8]. These models have several improvements over the OpenCLIP models, including an improved dataset [9], several innovations in the architecture and a larger text encoder. Additionally, during pre-training more augmentations are used, such as heavier random cropping, brightness/saturation jitter, and horizontal flip. For our comparison we use the ViT-B/16, ViT-L/14 & ViT-G/14 versions of the Perception Encoder Core models.

(iv) To investigate domain-specific optimization, we fine-tune the CLIPA-ViT-L/14 model using two strategies. Linear probing freezes the visual encoder and trains only the classification head, while full fine-tuning updates all model parameters. Both linear probing and full-finetuning is done for 80 epochs using AdamW [10] with a learning rate of 0.0005. For the first 5 epochs the backbone is frozen and afterwards has a learning rate multiplier of 0.005. Both approaches incorporate AutoAugment data augmentation [11] with random resizing to improve robustness to scale variations common in aerial and satellite imagery of military vehicles. These finetuned models test whether domain-specific training enhances performance beyond zero-shot capabilities.

**Occlusion Augmentations**

To evaluate CLIP's robustness under operational conditions, we implement two categories of synthetic occlusions: contiguous and dispersed patterns. Both categories are applied systematically from 0% to 100% occlusion in 5% increments. (Figure 2)

Contiguous occlusions represent scenarios where large connected regions are obscured. Slide blackout implements a vertical boundary progressing left-to-right across the image, simulating vehicles emerging from behind structures. Bar occlusion employs six evenly-spaced vertical bars with adjustable width, modeling observation through vegetation or regular obstacles.

Dispersed occlusions simulate fine-grained visual degradations. Random rain generates medium-sized vertical streaks with ±5° angle variation, mimicking precipitation effects on optical sensors. Random snow creates small circular masks (1-3 pixel diameter) distributed across the image, representing particulate interference. Grid dropout overlays a 10x10 grid on 224×224 pixel images with random cell selection, modeling systematic sensor failures or foliage.

In all occlusions it is guaranteed that when a certain percentage of occlusion is indicated that exact percentage of the image is obstructed. All occlusions use gray values (RGB: 128,128,128) to simulate partial information loss rather than complete blackout. The augmentation pipeline maintains consistent random seeds for reproducibility while preserving realistic variation across test samples.

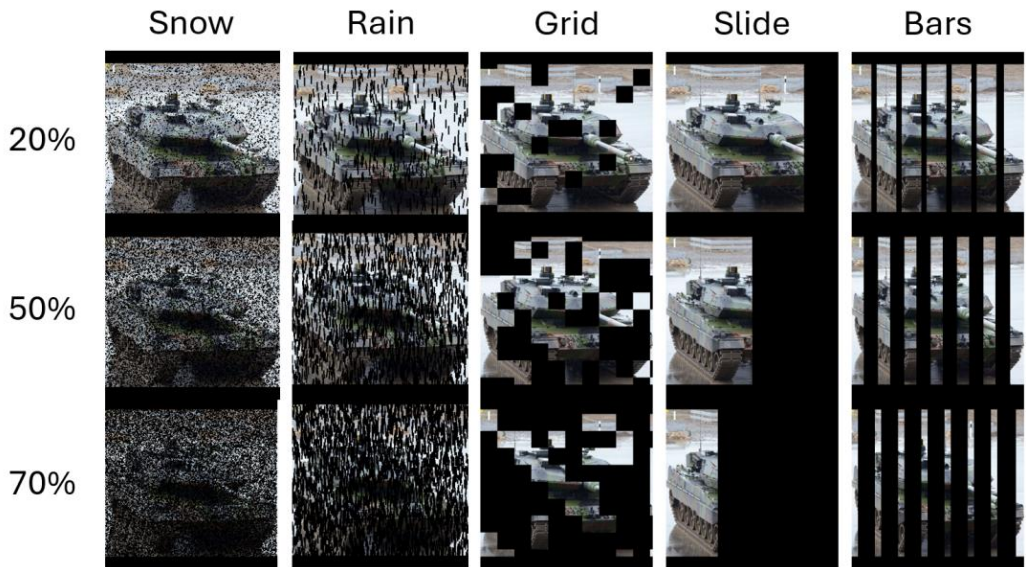

Figure 2: Example of increased occlusion using different types of augmentation

**Evaluation Metrics**

To quantitatively assess model robustness under various occlusion conditions, we employ two complementary metrics: standard accuracy and Normalized Area Under the Curve (NAUC).

For each experimental configuration, we compute classification accuracy under no occlusion, which is defined as: Accuracy = $\frac{N_{correct}}{N_{total}}$, where $N_{correct}$ represents the number of correctly classified military vehicle images and $N_{total}$ denotes the total number of test samples. While this metric provides interpretable performance measures at specific occlusion levels, direct comparison across multiple models and occlusion conditions requires extensive visualization (Figure 3).

To be able to do extensive model comparison and quantify overall robustness, we use the Normalized Area Under the Curve (NAUC) metric. This metric captures model performance degradation across the entire occlusion spectrum in a single scalar value. The NAUC is calculated as:

$$\text{NAUC} = \frac{1}{A_0} \cdot \frac{1}{n-1} \sum_{i=0}^{n-1} \frac{A_i + A_{i+1}}{2}$$

The $A_i$ represents the accuracy at occlusion level $i$ (in steps of 5%: 0%, 5%, 10%, ..., 100%), and the normalization by $A_0$ ensures that NAUC values are comparable across different models, regardless of their baseline performance. A NAUC value of 1.0 indicates perfect robustness (constant accuracy across all occlusion levels), while lower values indicate decreasing robustness to occlusion.

In all the experiments below we note $A_0$ as accuracy and the NAUC per occlusion type.

## EXPERIMENTS

We conducted a range of experiments to evaluate CLIP variants' robustness to occlusion on recognitions tasks of military vehicles. We compare between different architectures, different training regimes, different model sizes and evaluate whether occlusion resilience improves through model fine-tuning.

All experiments are evaluated in the same manner, using our 18-class military vehicle dataset. We apply five occlusion patterns; random snow, random rain, slide blackout, bars blackout, and grid dropout. Models are then evaluated using both classification accuracy ($A_0$) and Normalized Area Under the Curve metrics to capture performance degradation patterns. Each evaluation is run once.

Table 1: Performance comparison across model scales and training strategies. Models ordered by parameter count.

| Model | Parameter Count VE | Acc. | Dispersed Occl. | | Contiguous Occlusions | | | Average NAUC |
| --- | --- | --- | --- | --- | --- | --- | --- | --- |
| | | | NAUC Snow | NAUC Rain | NAUC Slide | NAUC Bars | NAUC Grid | |
| PE-Core-ViT-B/14-224 | 90M | 70.5% | 47.4% | 44.1% | 68.9% | 60.1% | 59.3% | 56.0% |
| CLIPA-ViT-L/14-224 | 304M | 55.6% | 54.0% | 54.6% | 76.0% | 71.9% | 69.0% | 65.1% |
| PE-Core-ViT-L/14-336 | 320M | 81.0% | 60.4% | 62.2% | **78.8%** | 81.1% | 76.7% | 71.8% |
| CLIPA-ViT-H/14-336 | 632M | 65.2% | 60.4% | 61.7% | 76.5% | 71.8% | 75.5% | 69.2% |
| CLIPA-ViT-G/14-336 | 1.85B | 68.5% | 64.6% | 65.0% | 75.3% | 80.0% | 75.2% | 72.0% |
| PE-Core-ViT-G/14-448 | 1.88B | **88.5%** | **71.4%** | **69.3%** | 78.6% | **83.7%** | **80.1%** | **76.6%** |

**Experiment 1: Occlusion types**

Our first experiment compares different occlusion types, both contiguous and dispersed. For this example we only utilize Table 1 with row PE-Core-ViT-L/14-336 (third row) displays the NAUC of the different occlusion types. These NAUC values are calculated from the accuracy values per occlusion percentage plotted in Figure .

Both dispersed occlusions (rain and snow) show a much lower NAUC score, 60.4% and 62.2%, respectively, than the contiguous occlusions (slide, bars and grid) with a NAUC of 78.8%, 81.1% and 76.7%, respectively. This indicates much higher robustness for the contiguous than for the dispersed occlusions.

The raw plot of the NAUC values as shown in Figure  shows the same trend. On all occlusion levels the dispersed occlusion types show lower accuracy on all occlusion levels with a growing gap in accuracy between 5% and 85% occlusion . Around 80% occlusion , bars occlusion lost 21.6 percentage point accuracy versus 60.2 percentage point accuracy for rain occlusion, which explains the large gap in NAUC score.

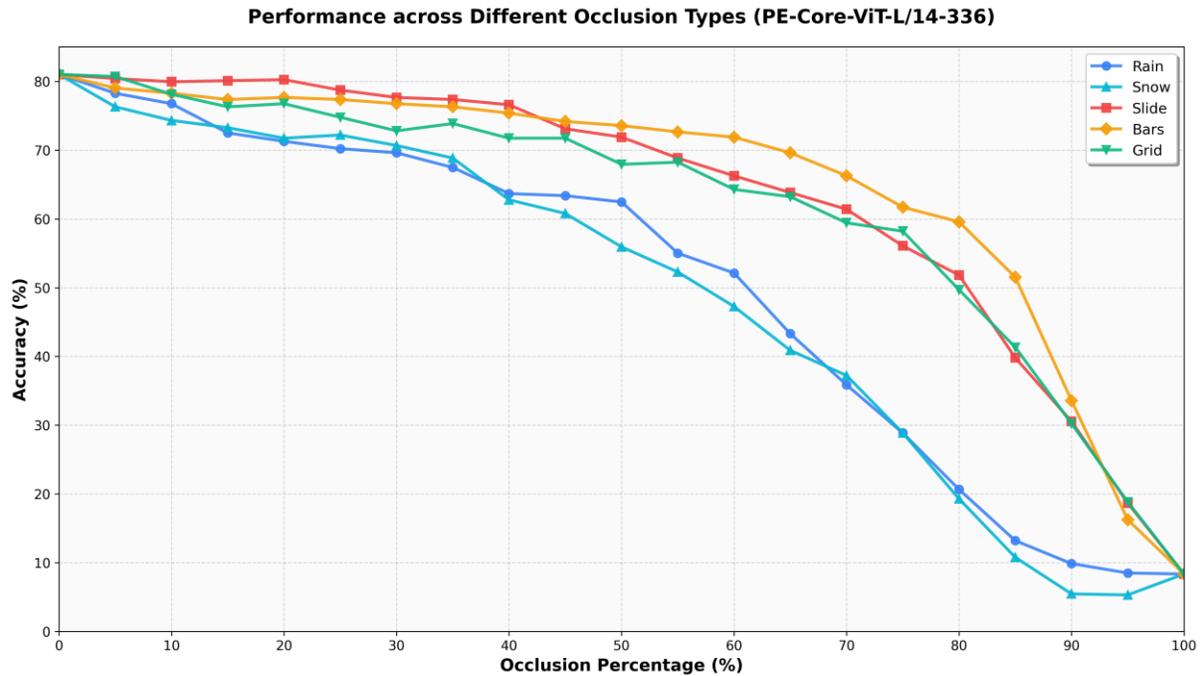

Figure 3: Example of accuracy over occlusion levels for different occlusion types.

**Experiment 2: Architectural Comparison**

Our second experiment isolates the effect of architectural design on occlusion robustness by comparing models with approximately equal parameter counts. We evaluate three CLIP variants: ResNet-50x4 (178M parameters) representing convolutional architectures, and two Vision Transformer configurations, ViT-B/32 (151M) and ViT-B/16 (150M), differing only in patch size. This controlled comparison reveals how fundamental architectural choices influence robustness to various occlusion patterns.

Table 2 presents zero-shot classification performance and NAUC scores across all occlusion types. Despite achieving identical baseline accuracy (34.8%) for both CLIP-RN50×4 and CLIP-ViT-B/16, the Vision Transformer architecture exhibits superior robustness to dispersed occlusions. Specifically, CLIP-ViT-B/16 achieves 60.0% NAUC on random snow conditions versus 38.2% for RN50×4, a 57% relative improvement. This performance gap is most pronounced for fine-grained occlusions (snow and rain), where ViT-B/16 outperforms the CNN by 21.8 and 22.1 percentage points respectively, demonstrating that architectural design impacts occlusion robustness.

Table 2: Architectural comparison of CLIP models with equivalent parameters.

| Model | Accuracy | Dispersed Occlusions | | Contiguous Occlusions | | | Average NAUC |
|---|---|---|---|---|---|---|---|
| | | NAUC Snow | NAUC Rain | NAUC Slide | NAUC Bars | NAUC Grid | |
| CLIP-RN50x4 | **34.8%** | 38.2% | 32.6% | 72.2% | 50.0% | 48.9% | 48.4% |
| CLIP-ViT-B/32 | 31.0% | 50.2% | 45.9% | **75.0%** | **60.6%** | 57.5% | 57.8% |
| CLIP-ViT-B/16 | **34.8%** | **60.0%** | **54.7%** | 74.8% | 54.5% | **64.5%** | **61.7%** |

Interestingly, all architectures demonstrate comparable robustness for most contiguous occlusions, with NAUC scores for slide blackout ranging from 72.2% to 75.0%. Which suggests that when large connected regions are occluded, the remaining visible portions provide sufficient information regardless of processing strategy. The ViT-B/16 model strikes an optimal balance, maintaining competitive accuracy (34.8%) while achieving the highest average NAUC (61.7%) across all occlusion types.

These findings establish that architectural choice significantly impacts occlusion robustness independent of parameter size, with Vision Transformers demonstrating clear advantages for military vehicle recognition under degraded visual conditions. However, it is important to note that, despite this overall improved robustness, performance degradation under dispersed occlusions remains evident for ViT-based models.

Examining the GradCAM visualizations for CLIP-ViT-B/16 under varying occlusion types and severities (see Figure 4) reveals notable differences in model attention. In the absence of occlusion, the heatmaps exhibit strong activation concentrated on the vehicle and its salient features. However, when dispersed occlusion (rain) is introduced, even at just 20% coverage, the model's attention begins to diffuse, partially shifting toward the background. At 70% dispersed occlusion, the majority of activation relocates away from the vehicle, suggesting a loss of robust feature localization.

Conversely, with contiguous (slide) occlusion, the GradCAM maps demonstrate that as large connected regions are occluded, the model's attention remains focused on the visible portions of the vehicle, effectively disregarding the occluded areas. This indicates a greater resilience in maintaining relevant spatial focus under contiguous occlusions compared to dispersed ones.

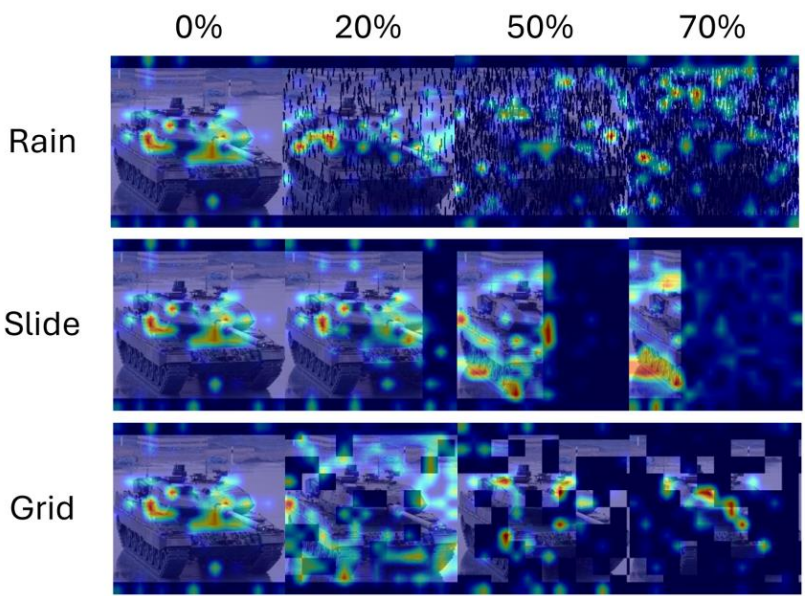

Figure 4: GradCAM for CLIP-ViT-B /16 under different occlusions

**Experiment 3: Model Scaling & Architectures**

The next experiment compares different models with increased parameter count and different training methodologies, whether these enhance robustness to visual degradations. Table 1 compares six state-of-the-art CLIP variants ranging from 90M to 1.88B parameters, including both CLIPA and PE-Core training methodologies. The results reveal distinct trends where model scale and training methodology independently influence occlusion robustness.

Model scaling demonstrates clear benefits for occlusion robustness, with both architectural families showing consistent improvements as parameter count increases. Within CLIPA models, scaling from 304M to 1.85B parameters improves average NAUC by 6.9 percentage points (65.1% to 72.0%), while PE-Core models achieve more substantial gains of 20.6 percentage points (56.0% to 76.6%) across the same scale range. The largest model, PE-Core-ViT-G/14-448, achieves the highest robustness across all metrics, demonstrating the combined benefits of architectural refinements and model scale. Comparing models with similar parameter counts reveals significant architectural advantages, with PE-Core-ViT-L/14-336 (320M) substantially outperforming CLIPA-ViT-L/14-224 (304M) by 25.5 percentage points in accuracy and 6.7 percentage points in average NAUC.

PE-Core models demonstrate greater resilience across all occlusion types, with PE-Core-ViT-G notably scoring: 69.3% rain NAUC compared to 65.0% for the comparable CLIPA model, and 80.1% grid dropout NAUC versus 75.2%. The training strategy remains consistent across different model scales, indicating that PE-Core's improved training methodology provide fundamental robustness benefits independent of parameter count. These results demonstrate that improved training methodology refinements and model scaling contribute independently to occlusion robustness, with PE-Core architectures delivering consistent performance gains across parameter ranges for military vehicle recognition under challenging operational conditions.

**Experiment 4: Domain Adaptation Effects**

The fourth experiment examines how domain-specific fine-tuning impacts classification accuracy and occlusion robustness. Table 3 compares zero-shot CLIPA-ViT-L/14 with two fine-tuning strategies: linear probing (training only the classification head) and full fine-tuning (updating all parameters). While both approaches substantially improve baseline accuracy, they reveal a critical limitation in achieving occlusion robustness through domain adaptation.

Table 3: Impact of fine-tuning strategies on accuracy-robustness trade-off.

| Model | Training | Acc. | Dispersed Occl. | | Contiguous Occlusions | | | Average NAUC |
| --- | --- | --- | --- | --- | --- | --- | --- | --- |
| | | | NAUC Snow | NAUC Rain | NAUC Slide | NAUC Bars | NAUC Grid | |
| CLIPA-ViT-L/14 | Zero-shot | 55.6% | **54.0%** | **54.6%** | 76.0% | 71.9% | 69.0% | **65.1%** |
| CLIPA-ViT-L/14 | Linear Probe | 84.0% | 36.0% | 42.5% | 73.9% | 68.8% | 67.4% | 57.7% |
| CLIPA-ViT-L/14 | Full FT | **93.3%** | 44.5% | 49.4% | 75.4% | **74.3%** | **71.1%** | 62.9% |

Linear probing achieves 84.0% accuracy, a 28.4 percentage point improvement, but dramatically reduces robustness to dispersed occlusions, with snow NAUC dropping from 54.0% to 36.0%. This 33% relative decrease suggests that adapting only high-level features negatively impacts robustness against low-level visual degradations. Full fine-tuning partially mitigates this issue, reaching 93.3% accuracy while maintaining better robustness (44.5% snow NAUC), though still below zero-shot performance for dispersed occlusions.

Detailed analysis of per-occlusion-level accuracy reveals more concerning patterns for linear probing performance. As shown in Figure 5, linear probing accuracy degrades rapidly, when occlusion is introduced. At 30% occlusion the linear probing model (43.1%) performs even worse than the zero-shot model (49.1%). This sharp decline in performance when occlusion is introduced suggests overfitting to clean image features that become unreliable under occlusion. In contrast, full fine-tuning always has better accuracy or at least on par with zero-shot over all occlusion levels (Figure 5).

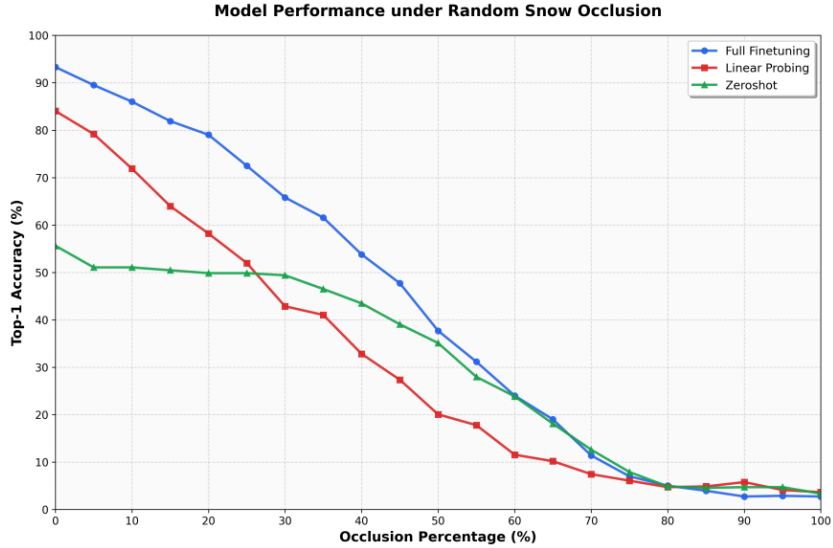

Figure 5: Performance on snow occlusion for finetuned models

## DISCUSSION

Our experiments reveal several critical insights into the robustness of CLIP models under occlusion, highlighting architectural, methodological, and training-related factors.

Our first experiment compares the difference in contiguous and dispersed occlusions. We already noted that larger amounts of occlusion lead to increased accuracy gaps (Figure 3). The tables in experiment 2 and experiment 3 reveal a continuation of this trend and show that NAUC is higher for contiguous occlusions than for dispersed occlusions in all experiments. In Figure 4, we can qualitatively observe these effects. The first column of the GradCAM images show a clear focus on important features of the main battlefield tank. With increased occlusion we notice a confusion of focus in the dispersed occlusion samples (grid and rain) and already from 20% occlusion a confusion between foreground and background. However, for the contiguous occlusion (slide) we can still see a focus on the main features while ignoring the occluded area. Combined, these results suggest that dispersed occlusions are inherently more difficult than contiguous occlusions.

In the second experiment, we concluded that Transformer-based CLIP models and CNN-based models achieved similar accuracies on non-occluded samples. However, when we evaluated robustness under dispersed occlusions, Transformer models displayed substantially stronger performance. This can be explained by their architectural differences. CNNs tend to perform worse under dispersed occlusions due to their reliance on local receptive fields and hierarchical spatial processing. When occlusions corrupt a significant portion of a local receptive field, the resulting feature maps become corrupted, and the network struggles to recover meaningful representations. Although Transformers also initially process local patches, they immediately apply a global attention mechanism that allows them to consider relationships across the entire image. This enables them to dynamically balance which features are relevant and which are likely occluded or noisy. In contrast, CNNs only achieve a global view at the deeper layers of the network, where it becomes much harder to disentangle which features were affected by occlusion and which remain informative. This architectural difference gives Transformers a distinct advantage in robustness when vehicles are occluded.

Lastly, experiments 3 and 4 indicate that occlusion robustness is shaped not only by architecture but also by training methodology. In experiment 3 (Table 3), the PE-Core model—despite sharing the same architecture and input size as CLIPA—achieves higher robustness under occlusion, driven mainly by stronger augmentations (cropping, color jittering) and an improved dataset rather than architectural changes. In experiment 4, linear probing exhibits markedly poorer robustness than zero-shot (partly due to NAUC's normalization to a higher clean-accuracy ceiling and as visualized in Figure 5), while full finetuning recovers part of this degradation. The experiment suggests that occlusion robustness primarily arises in the early and middle layers of the visual backbone where patch-level features are formed. By the time the classification head operates on high-level semantics, crucial spatial cues may already have been discarded, and linear probing cannot recover them—even if it improves clean accuracy.

Taken together, these experiments show that training strategy has substantial leverage. Generic augmentations help but do not sufficiently address high-occlusion regimes. We therefore recommend (i) **targeted augmentations** that mimic real occlusion patterns (e.g., grid masks [12], rain streaks [13], sliding blocks), (ii) **masking-based training** such as masked image modeling or dynamic masking during finetuning to encourage global context awareness [14], and (iii) **occlusion-aware loss objectives** to reduce sensitivity to corrupted regions [15]. Treating occlusion as a core training objective rather than a nuisance variable can significantly improve robustness beyond what generic augmentations achieve.

## CONCLUSIONS

This work systematically evaluated the robustness of CLIP-based vision-language models to occlusion in military vehicle classification, addressing a critical gap between model training conditions and operational realities. Using a custom dataset of 18 military vehicle classes and five occlusion types, we assessed performance across architectures, model scales, and fine-tuning strategies using both accuracy and Normalized Area Under the Curve (NAUC) metrics.

Four key findings emerged. (i) Occlusion type matters: dispersed occlusions such as snow and rain cause significantly greater performance degradation than contiguous occlusions, underscoring the need for occlusion-aware evaluation protocols. (ii) Architectural design strongly influences robustness: Transformer-based CLIP models consistently outperform CNN-based counterparts under dispersed occlusions, confirming the advantage of global attention mechanisms for handling fragmented visual information. (iii) Model scaling and improved training methodologies (e.g., PE-Core) independently enhance robustness, but scaling alone does not eliminate vulnerabilities to fine-grained occlusions. (iv) Domain-specific fine-tuning improves clean-image accuracy but introduces a trade-off with robustness, particularly when adaptation is limited to the classification head. Full fine-tuning mitigates this effect but still falls short of zero-shot robustness due to the definition of robustness.

These findings highlight that robustness cannot be assumed from baseline accuracy or model size alone. To ensure reliable deployment in defense applications, occlusion robustness must be treated as a primary design objective. Future research should explore targeted occlusion augmentations, masking-based training strategies, and occlusion-aware loss functions to strengthen resilience under operational conditions. By integrating these strategies, vision-language models like CLIP can move closer to meeting the stringent requirements of real-world military environments where partial visibility is the norm.

## ACKNOWLEDGMENTS

This work received funding from the European Defence Fund (EDF) through the projects FaRADAI (Frugal And Robust AI for Defence Advanced Intelligence), grant agreement № 101103386; and STORE (Shared daTabase for Optronics image Recognition and Evaluation), grant agreement № 101121405.